\documentclass[10pt,twocolumn,letterpaper]{article}

\usepackage{cvpr}              

\usepackage{graphicx}
\usepackage{amsmath}
\usepackage{amssymb}
\usepackage{booktabs}

\usepackage[pagebackref,breaklinks,colorlinks]{hyperref}

\usepackage[capitalize]{cleveref}
\crefname{section}{Sec.}{Secs.}
\Crefname{section}{Section}{Sections}
\Crefname{table}{Table}{Tables}
\crefname{table}{Tab.}{Tabs.}

\def\cvprPaperID{3} 
\def\confName{CVPRW}
\def\confYear{2023}

\newcommand{\mysection}[1]{\vspace{2pt}\noindent\textbf{#1}}
\newcommand{\Md}[1]{$\mathcal{M}_{data}^{#1\%}$}
\newcommand{\Mp}[1]{$\mathcal{M}_{perf}^{#1\%}$}

\newcommand\blfootnote[1]{%
  \begingroup
  \renewcommand\thefootnote{}\footnote{#1}%
  \addtocounter{footnote}{-1}%
  \endgroup
}
\begin{document}

\title{Towards Active Learning for Action Spotting in Association Football Videos}

\author{Silvio Giancola$^{*,1}$ 
\quad
Anthony Cioppa$^{*,1,2}$  
\quad
Julia Georgieva$^{3}$  
\quad
Johsan Billingham$^{4}$  
\and
Andreas Serner$^{5}$  
\quad
Kerry Peek$^{6}$  
\quad
Bernard Ghanem$^{1}$  
\quad
Marc Van Droogenbroeck$^{2}$  
\and  {\small $^1$IVUL, KAUST}
\quad {\small $^2$TELIM, University of Liège }
\quad {\small $^3$Curtin School of Allied Health, Curtin University }
\and {\small $^4$Football Technology \& Innovation, FIFA }
\quad {\small $^5$FIFA Medical, FIFA }
\quad {\small $^6$Faculty of Medicine and Health, The University of Sydney }
}






\maketitle

\begin{abstract}
Association football is a complex and dynamic sport, with numerous actions occurring simultaneously in each game. Analyzing football videos is challenging and requires identifying subtle and diverse spatio-temporal patterns. Despite recent advances in computer vision, current algorithms still face significant challenges when learning from limited annotated data, lowering their performance in detecting these patterns. In this paper, we propose an active learning framework that selects the most informative video samples to be annotated next, thus drastically reducing the annotation effort and accelerating the training of action spotting models to reach the highest accuracy at a faster pace. Our approach leverages the notion of uncertainty sampling to select the most challenging video clips to train on next, hastening the learning process of the algorithm. We demonstrate that our proposed active learning framework effectively reduces the required training data for accurate action spotting in football videos. We achieve similar performances for action spotting with NetVLAD++ on SoccerNet-v2, using only one-third of the dataset, indicating significant capabilities for reducing annotation time and improving data efficiency. We further validate our approach on two new datasets that focus on temporally localizing actions of headers and passes, proving its effectiveness across different action semantics in football. We believe our active learning framework for action spotting would support further applications of action spotting algorithms and accelerate annotation campaigns in the sports domain.

\blfootnote{\textbf{(*)} Denotes equal contributions \\Contacts: silvio.giancola@kaust.edu.sa, anthony.cioppa@uliege.be. 
}

\end{abstract}

\begin{figure}[t]
    \centering
    \includegraphics[width=\linewidth,trim={0 0 0 20mm},clip]{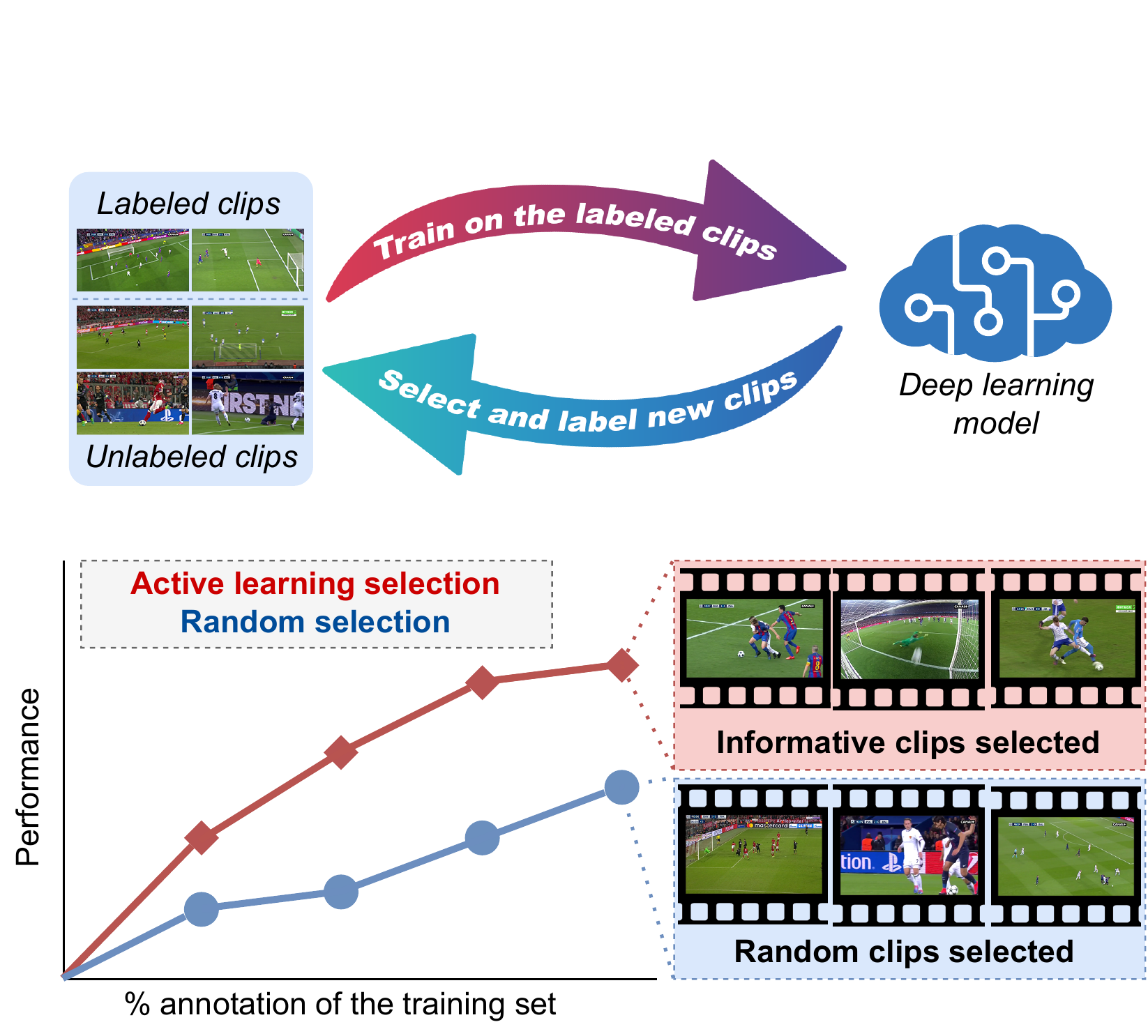}
    \caption{
    \textbf{Active learning for action spotting.}
    Given a video clip dataset, our active learning framework iteratively 
    \textbf{(i)} trains a deep learning model on the labeled clips, and 
    \textbf{(ii)} selects the next video clips to be labeled by an oracle.
    By actively selecting the most informative video samples to annotate next, we accelerate the tagging of unlabeled datasets for training action spotting models.
    }
    \label{fig:graphical_abstract}
\end{figure}

\section{Introduction}
\label{sec:intro}

Video analysis is a rapidly evolving field with numerous applications in various domains, such as surveillance~\cite{Sreenu2019Intelligent}, sports~\cite{Thomas2017Computer}, and autonomous driving~\cite{Pavel2022VisionBased}. One of the essential tasks in video analysis is action spotting, which aims at identifying and precisely localizing specific actions anchored with a single timestamp in video sequences. This task has gained considerable attention in recent years due to its importance in various applications, 
such as video search~\cite{Castanon2016Exploratory,Soldan2021MADAS}, video summarization~\cite{Cioppa2020AContextAware,Ekin2003Automatic}, and activity recognition~\cite{Caba2015ActivityNet,Host2022AnOverview}.

Traditionally, action spotting is addressed using supervised learning techniques, where a labeled dataset is used to train a classifier that can recognize actions and temporally localize them in the videos. However, annotating large-scale video datasets is time-consuming and expensive, limiting the scalability and applicability of supervised learning approaches. Active learning comes as a promising approach that can mitigate the need for completely annotating large datasets by selecting the most informative samples needed for labeling.

In this paper, we propose an active learning framework for action spotting in association football videos, which aims to reduce the annotation effort and improve the overall performance of the system with respect to the number of annotations.
Our framework, illustrated in Figure~\ref{fig:graphical_abstract}, integrates active learning with state-of-the-art action spotting methods, where the labeled set is iteratively expanded with informative samples selected from an unlabeled dataset. We evaluate our approach on several benchmark datasets and compare it with a naive random selection approach that does not leverage active learning. We also analyze the impact of different query strategies on the performance of the system.


\mysection{Contributions.} Our contributions may be summarized as follows:
\textbf{(i)} 
We propose the first active learning framework for the task of action spotting that iteratively selects relevant clips to be annotated next.
\textbf{(ii)} 
We compare a couple of active learning selection strategies based on uncertainty sampling on several benchmark datasets with state-of-the-art action spotting methods.
\textbf{(iii)} 
We provide a comprehensive analysis showing the capability of our framework to significantly reduce the quantity of annotation needed to reach desired performances.

\section{Related work}
\label{sec:relatedwork}


\subsection{Sports video understanding}

The challenging and nuanced nature of sports video analysis has made it an increasingly popular research topic in recent years~\cite{Moeslund2014Computer, Thomas2017Computer}.
The availability of large-scale datasets has played a crucial role in enabling progress in these tasks. Examples of such large-scale datasets include those developed by Pappalardo~\etal~\cite{Pappalardo2019APublic}, Yu~\etal~\cite{Yu2018Comprehensive}, SoccerDB~\cite{Jiang2020SoccerDB}, SoccerTrack~\cite{Scott2022SoccerTrack}, and DeepSportRadar~\cite{VanZandycke2022DeepSportradarv1}.
The SoccerNet dataset, introduced by Giancola~\etal~\cite{Giancola2018SoccerNet}, has become the most comprehensive resource for labeled data related to video understanding in football. It includes benchmarks for ten different tasks, such as action spotting~\cite{Deliege2021SoccerNetv2}, camera calibration~\cite{Cioppa2022Scaling}, and player tracking~\cite{Cioppa2022SoccerNetTracking}.

Lately, deep learning-based methods have become the go-to approach for many sports video analysis tasks, thanks to their remarkable performance and ability to extract high-level features from raw data. For instance, automatic methods based on deep learning have shown impressive results in tasks such as player tracking~\cite{Maglo2022Efficient} and re-identification with occlusion~\cite{Somers2023Body}, 3D shuttle trajectory reconstruction in badminton~\cite{Liu2022MonoTrack}, medical risk assessment in rugby~\cite{nonaka2022end}, tactics analysis~\cite{Suzuki2019Team}, pass feasibility~\cite{ArbuesSanguesa2020Using}, and talent scouting~\cite{Decroos2019Actions}. Additionally, deep learning-based methods have also allowed researchers to leverage large-scale datasets effectively. Despite their successes, deep learning-based methods still face several challenges, such as dealing with noisy and incomplete data, accounting for complex game scenarios, and generalizing to new domains. As such, there is still ample room for improvement and research in the field of sports video understanding.
Furthermore, annotating large-scale datasets for sports video analysis tasks is a time-consuming and expensive process, requiring significant human resources and expertise. As a solution, Vandeghen~\etal~\cite{Vandeghen2022SemiSupervised} proposed a semi-supervised method for player detection, leveraging a large unlabeled dataset. Vats~\etal~\cite{Vats2022IceHockey} propose a weakly supervised approach for player identification using transformers. To address this data issue, we propose an active learning approach for action spotting in football that aims at reducing the amount of annotated data needed while maintaining high task performance.

\subsection{Action spotting}

Action spotting is an important task in football video understanding, as it involves localizing specific events in an untrimmed football broadcast video, including, for instance, penalties, goals, or corners. Unlike temporal activity localization~\cite{Caba2015ActivityNet}, action spotting describes events using a single timestamp, following the definition of actions defined in the football rules~\cite{IFAB2022Laws}. 
Recent studies have explored the use of large-scale datasets such as SoccerNet~\cite{Giancola2018SoccerNet}, which has been expanded from $3$ to $17$ classes to encompass all possible actions that occur during a game~\cite{Deliege2021SoccerNetv2}. This dataset has generated significant interest in the research community, as evidenced by the open challenges~\cite{Giancola2022SoccerNet}, showing that action spotting is currently experiencing a high level of activity and attention in the research and industrial communities.

The first method for action spotting was proposed by Giancola~\etal~\cite{Giancola2018SoccerNet}, which is based on temporal pooling. Later, they refined their method by aggregating the temporal context~\cite{Giancola2021Temporally}. Rongved~\etal~\cite{Rongved2021Automated} proposed an approach based on applying a 3D ResNet directly to the video frames in a 5-second sliding window fashion. 
Vanderplaetse~\etal~\cite{Vanderplaetse2020Improved} and Xarles~\etal~\cite{Giancola2022SoccerNet} combined visual and audio features in a multimodal approach. Cioppa~\etal~\cite{Cioppa2020AContextAware} introduced a context-aware loss function to model the temporal context surrounding the actions. Vats~\etal~\cite{Vats2020Event-arxiv} used a multi-tower CNN that accounts for the uncertainty of the action locations, and Tomei~\etal~\cite{Tomei2021RMSNet} fine-tuned a feature extractor and used a masking strategy to focus on the frames after the actions.

The current state-of-the-art on SoccerNet-v2 is held by Soares~\etal~\cite{Soares2022Temporally,Soares2022Action-arxiv}, who won the SoccerNet 2022 challenge by proposing an anchor-based approach. They define an anchor as a pair formed by a time instant and an action class, with time instants sampled densely. For each anchor, both a detection confidence and a fine-grained temporal displacement are inferred, with the displacement indicating exactly when an action is predicted to happen. Their approach results in a substantial improvement in temporal precision. 
Hong~\etal~\cite{Hong2022Spotting-arxiv}, the runner-ups of the 2022 challenge, proposed the first precise temporal spotting (PTS) method where both the feature extractor and the spotting head are trained in an end-to-end fashion. They rely on a light-weight RegNet architecture, including a GSM~\cite{Sudhakaran2020GateShift} module and a GRU~\cite{Cho2014OnThe} module on top that classifies each frame into an action class or background. Other methods also focused on spatio-temporal encoders~\cite{Darwish2022STE}, graph-based architecture~\cite{Cartas2022AGraphbased}, or transformer architectures~\cite{Zhu2022ATransformerbased}.

Despite their impressive performance, all state-of-the-art methods rely on supervised learning, which requires a large-scale annotated dataset. However, in sports video analysis, the actions to spot may change over time, which would require re-annotating the dataset. In this work, we study how to efficiently re-annotate such datasets with active learning techniques, which aim at selecting the most informative samples for annotation, thereby minimizing the annotation effort while maintaining high task performance.

\subsection{Active learning}

Active Learning has been successfully applied in a wide range of applications, including image understanding~\cite{gal2017deep,houlsby2011bayesian}, 
video understanding~\cite{heilbron2018annotate}, 
natural language processing~\cite{thompson1999active},
speech recognition~\cite{hakkani2002active}, and
chemistry~\cite{duros2017human}.
The main objective of active learning is to select the most informative unlabeled samples for annotation and use the minimal amount of label data  to achieve specific performance. 
The main strategies for active learning include 
\textit{uncertainty sampling}~\cite{lewis1994heterogeneous,tong2001support,joshi2009multi}, 
\textit{diversity maximization}~\cite{yang2015multi,sener2017active},
\textit{query-by-committee}~\cite{seung1992query,freund1997selective,gilad2005query,iglesias2011combining}, and
\textit{expected error}~\cite{hoi2006batch,joshi2012scalable,konyushkova2017learning,yoo2019learning}.
We refer to \cite{settles2009active,Ren2021ASurvey} for a comprehensive and more generic literature review on active learning.

\mysection{Uncertainty sampling.} 
Those methods sample the unlabeled data that confuses most of the action spotting models trained thus far.
Tong~\etal~\cite{tong2001support} proposed the use of a support vector machine algorithm for conducting effective relevance feedback for image retrieval.
The active learning method introduced by Joshi~\etal~\cite{joshi2009multi} selects unlabeled data that the model finds hardest to classify. The selection is based on the entropy of the output of the classifier or a ``Best versus Second Best'' (BvSB) paradigm.

\mysection{Diversity maximization.}
Those active learning approaches select samples that best represent the whole space of the available unlabeled set.
Yang~\etal~\cite{yang2015multi} proposed a method that maximizes the diversity of the samples. They investigated this approach on diverse visual recognition tasks, including action recognition, object classification, scene recognition, and event detection.
Similarly, Sener~\etal~\cite{sener2017active} modeled the selection process as a core-set problem. They sample representative subsets of images by minimizing the L2 distance with the remaining samples in the dataset.

\mysection{Query-by-committee.}
In this paradigm, the next batch to annotate is chosen according to the principle of maximal disagreement between a committee of student algorithms trained on the same labeled dataset.
Seung~\etal~\cite{seung1992query} introduced this approach, further analyzed in a Bayesian framework by Freund~\etal~\cite{freund1997selective}.
Houlsby~\etal~\cite{houlsby2011bayesian} further investigate the relationship of Query-by-Committee with information gain theory.

\mysection{Expected error.}
Those methods attempt to learn a metric that correlates with the error of classifying specific samples.
Learning Active Learning (LAL)~\cite{konyushkova2017learning} learns to regress the error reduction for a candidate sample. The active learning scores are learned in a supervised fashion on the error loss in the training dataset.
Similarly, Yoo~\etal~\cite{yoo2019learning} also learn a ``loss prediction module'' agnostic to any task. By doing so, they actively select samples with higher predicted loss, expecting those samples to provide significant novel information to minimize for on the next train step.

\mysection{Active learning for temporal video analysis.}
While active learning has been extensively analyzed on generic setups, only a few works apply those approaches to temporal video analysis.
Brandla~\etal~\cite{bandla2013active} proposed an active learning method for temporal activity localization (TAL) algorithms, based on \textit{uncertainty sampling}~\cite{lewis1994heterogeneous}. 
Heilbron~\etal~\cite{heilbron2018annotate} further investigated active learning in TAL with an empirical study of different active learning paradigms, with LAL~\cite{konyushkova2017learning} performing best.

Following the previous literature, we formalize the first active learning workflow for action spotting. We analyze a couple of \textit{uncertainty sampling} methods and set the ground for more active learning approaches for action spotting.

\section{Active learning for action spotting}
\label{sec:method}

\begin{figure*}[t]
    \centering
    \includegraphics[width=\linewidth]{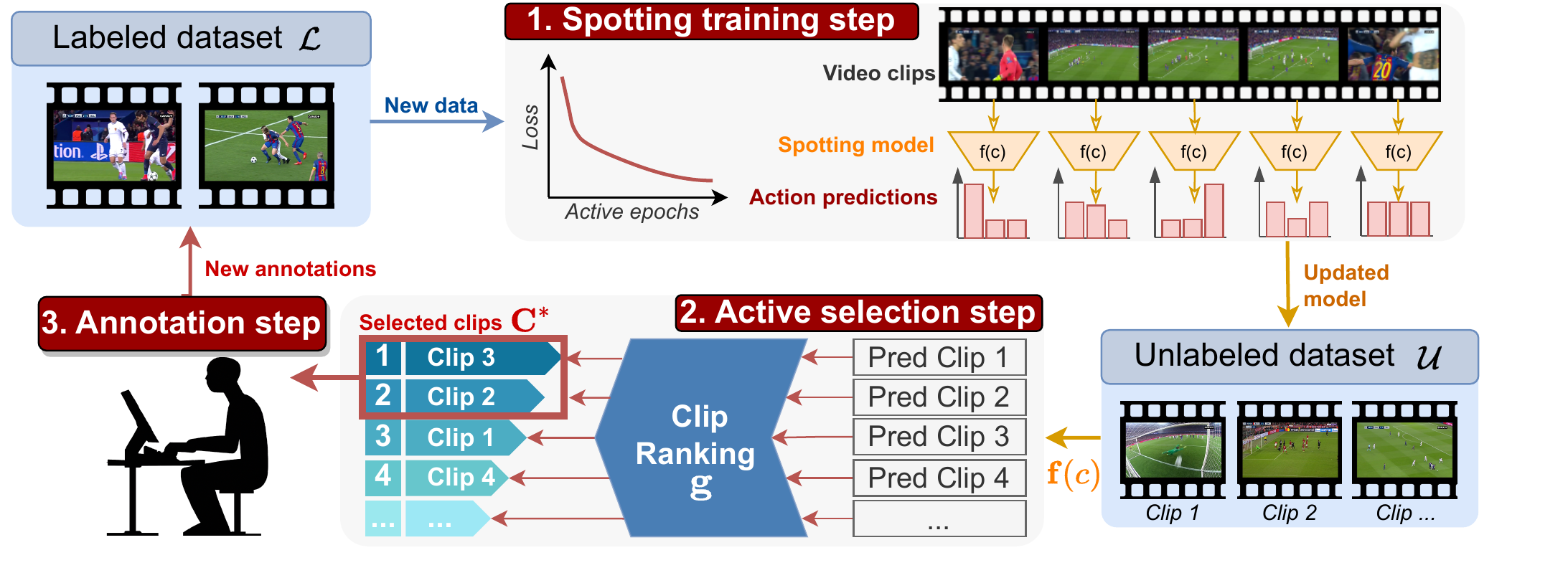}
    \caption{
    \textbf{Active learning pipeline for action spotting.}
    We start from a small labeled dataset $\mathcal{L}$ on which we train an action spotting model whose inference function is denoted $f$.
    With the trained model, we select from an unlabeled dataset $\mathcal{U}$ which sample to annotate next. For that, we first collect the prediction of the model $f(\mathcal{U})$ for each clip and pass the predictions through our selection function $g$ that ranks the clips to select $\mathbf{C}^*$. All selected clips are then passed to the oracle (human annotator) to provide both the class and localization of all actions within that clip. These new annotated data are then added to the labeled dataset and used for the next training iteration. The process is repeated iteratively until the desired performance is reached or the unlabeled dataset is empty.
    }
    \label{fig:pipeline}
\end{figure*}

We propose the first active learning framework for the task of action spotting. 
Our framework aims at training an accurate action spotting model using a minimal amount of labeled data.
Following the literature on \textit{uncertainty sampling} active learning, we identify three key steps to achieve this objective: 
\textbf{(1)}~Train an action spotting model on a labeled dataset that grows at each active learning step. 
\textbf{(2)}~Select the most informative data from an unlabeled pool using an active learning algorithm. 
\textbf{(3)}~Label the selected clips by an oracle and include the new data and annotations in the labeled set.
An overview of our complete framework is depicted in Figure~\ref{fig:pipeline}.

Formally, given a video $v$, the task of action spotting is to identify all action spots $\mathbf{S}=\{s_1,...,s_M\}$ inside that video. 
A spot $s_m$ comprises the action class (\eg penalty, goal, etc.) and a temporal anchor.
At each active learning step $\tau$, an action spotting model, whose inference function is denoted $f_\tau$, is trained using a set of labeled samples $\mathcal{L}_\tau$. The model details and training procedure are described in Section~\ref{subsec:training}.
Then, an active learning algorithm $g$ selects an optimal set of unlabeled samples $\mathbf{C}^*$ from a pool $\mathcal{U}_\tau$. Several active learning algorithms are presented in Section~\ref{subsec:selection}.
Subsequently, an oracle, described in Section~\ref{subsec:annotation}, provides the ground-truth annotations of the actions spots (\ie action classes $k$ and temporal anchors $t$) for the selected samples $\mathbf{C}^*$.
The labeled set $\mathcal{L}_\tau$ is then augmented with these newly labeled clip instances $\mathbf{C}^*$ following $\mathcal{L}_{\tau+1}=\mathcal{L}_{\tau} \cup \mathbf{C}^*$. 
This process is repeated until the model reaches a desired performance or the set $\mathcal{U}_\tau$ is exhausted. 
Since the most expensive step consists in labeling the samples, the objective of our framework is to minimize the number of times the oracle is queried by proposing an efficient active selection algorithm.


\subsection{Model training step}
\label{subsec:training}

\mysection{Datasets.}
The datasets for action spotting typically consist of a list of $L$ untrimmed videos $\mathbf{V}=\{v_1,...,v_L\}$, each video being annotated with a set $\mathbf{S}$ of action spots $s=\{k,t\}$ of class $k$ among $K$ classes, anchored with a single timestamp $t$. 
Since training action spotting models on long untrimmed video is not yet possible due to hardware constraints (\eg GPU memory or computation time), they are typically trained on clips extracted from the video. In this work, we consider that each video $v_l$ can be viewed as a set of $N$ fixed-length non-overlapping clips $\mathbf{C_l}=\{c_l^1,...,c_l^N\}$. 
Each clip $c_l^n$ can be annotated with a list of temporally-anchored action spots $\mathbf{S_l^n}=\{s_{l,1}^n,...,s_{l,M}^n\}$.

\mysection{Video encoder.}
Typical action spotting models are composed of a video encoder $\mathbf{H}$ followed by an action spotting head $\mathbf{A}$.
Given a trimmed video clips $c_l^n$ composed of $J$  frames, a video encoder extracts a compact features representation $\mathbf{H(c_l^n)}=\{h_{l,1}^n,...,h_{l,J}^n\}$ for each frame.
This frame feature encoder is usually pre-trained on an external dataset and then either frozen or fine-tuned on the action spotting dataset.
Due to the diversity of features dimension, it is common to homogenize their dimensionality with PCA to produce an even more compact and standardized frame representation.
Also, these feature encoders can either be applied on the entire video clip, leveraging the temporal information, or independently on each frame of the clip, commonly requiring less computational power and memory.
A typical choice for action spotting baselines is to extract frame features with a learnable CNN-based encoder such as the frame-based ResNet encoder, or the video-based I3D/C3D encoders~\cite{Giancola2018SoccerNet}. 

\mysection{Action spotting head.}
Given a set of compact frame features representation $\mathbf{H(c_l^n)}=\{h_{l,1}^n,...,h_{l,J}^n\}$, an action spotting head $\mathbf{A}$ temporally combines the descriptors and outputs a list of predicted action spots $\mathbf{\hat{S}_l^n} = \mathbf{A(H(c_l^n))}= \{\hat{s}_{l,1}^n,...,\hat{s}_{l,M'}^n\}$ for the current clip $c_l^n$.
This list of predictions can be obtained in two ways. A first category of action spotting models~\cite{Cioppa2020AContextAware} directly regresses the predicted location and class.
In this work, we focus on a second category, that first outputs $K+1$ class scores (including the background) per frame or per clip. The exact localizations $t$ of the actions are then extracted using a non-maxima-suppression algorithm on the predicted class scores over time.
The complete mathematical function of action spotting methods $f_\tau$ can therefore be expressed as
$\mathbf{A} \circ \mathbf{H}$.




\mysection{Training.}
We define the labeled dataset $\mathcal{L}_\tau$ at the active learning step $n$ of size $|\mathcal{L}_\tau|$ as 
$\mathcal{L}_\tau=\{(c^{train}_1,  \mathbf{S}_1),$$ ..., (c^{train}_{|\mathcal{L}_\tau|}, \mathbf{S}_{|\mathcal{L}_\tau|})\}$.
At each active learning step, the action spotting baseline is trained on $\mathcal{L}_\tau$.
In our framework, we consider several training paradigms.
The first one consists in training the action spotting module from scratch at each active learning step. 
This training may be done until convergence or on a particular number of epochs. 
The advantage is that the deep learning model usually trains better as it. 
However, the drawback is that it may require a lot of time to train each epoch.
A faster training paradigm consists in fine-tuning the model obtained at the previous active learning step, either until convergence or for a fixed number of epochs.
This reduces the training time but does not ensure convergence. 
For instance, if the network diverges during the first steps due to the low amount of training data, it may be unable to recover later on.
We study these training paradigms in the experimental section.


\mysection{Inference.}
At test time, the model produces the predictions over a full video while it has been trained only on clips. 
One common way to solve this mismatch is to split the video into overlapping or non-overlapping clips.
Each clip is processed independently and the results are aggregated along the video.
The spotting performances are evaluated using the mean average precision (mAP) from successfully spotting an action within a given temporal tolerance $\delta$.
The main associated metric is the Avg-mAP, where the mAP are averaged for various values of $\delta$-tolerance between ground truth and predicted action spots.
We use the typical metrics~\cite{Giancola2022SoccerNet} \emph{tight} Avg-mAP (with $\delta$ ranging from $1$ to $5$ seconds) and \emph{loose} Avg-mAP (with $\delta$ ranging from $5$ to $60$ seconds).



\subsection{Active selection step}
\label{subsec:selection}

The next step consists in selecting clips from the unlabeled dataset, defined as a set of unlabeled video clips $\mathcal{U}_\tau=\{c^{u}_1, ..., c^{u}_{|\mathcal{U}_\tau|}\}$ of size $|\mathcal{U}_\tau|$. 
The objective of the active selection step is to create a function $g$ that selects a new set of clips $\mathbf{C}^*$ from $\mathcal{U}_\tau$. The main challenge is to ensure that the function chooses samples that are likely to have the greatest impact on improving the action spotting model.
As described in Section~\ref{sec:relatedwork}, there exists many active learning workflows. In this work, we focus on the particular case of \textit{uncertainty sampling}.
%
%
In particular, we analyze the predictions of the action spotting model trained at the previous active learning step.
The predictions are a set of $(K+1)$ probability values for each class, either per frame or per clip.
In the case of black-box models, the class confidence scores are the sole uncertainty information returned by a prediction.
Following the literature on active learning for image classification, we construct two selectors leveraging the Uncertainty Measure (UM) and the Entropy Measure (EM).
In the following, we show how to implement these methods for action spotting predictions.

\mysection{Uncertainty measure.}
The Uncertainty Measure (UM) solely considers the confidence scores associated with each clip or frame. 
Given a confidence score $p_k$, the active learning score is inversely proportional to its distance to a confused confidence of $0.5$.
The Entropy Measure is formally defined as follows: 
\begin{equation}
    \mbox{UM}=1-2\times |p_k - 0.5| \, .
    \label{eq:UM}
\end{equation}
In the particular case of action spotting, this score is computed per frame and then averaged or max pooled over all frames.


\mysection{Entropy measure.}
The Entropy Measure (EM) considers the distribution of the confidence for all the classes.
Such estimation requires access to the confidence score for all classes, which might not be accessible in the case of black-box algorithms, that only returns the highest confidence for the selected class.
Based on the list of confidence scores {$p_1,...,p_K$}, we extract an active learning score inversely proportional to the uniformity of the distribution between the predictions.
The Entropy Measure is formally defined as: 
\begin{equation}
    \mbox{EM}=-\sum_{i=1}^K p_i \log(p_i)\, .
    \label{eq:EM}
\end{equation}

\mysection{Selecting samples.}
We leverage the function $g$ to select the top-k most informative clips $\mathbf{C}^*$ with the highest active score.
In this work, we study several approaches to select the number of clips $|\mathbf{C}^*|$ at each active learning step.
A first approach consists in selecting a fixed number of clips at each active learning step.
A second approach consists in selecting an increasing number of clips.
This has the advantage of selecting only relevant clips at the beginning, when the model still requires highly informative clips.



\subsection{Annotation step}
\label{subsec:annotation}

Once the clips have been selected by the active learning step, they need to be annotated by an oracle (which is a human annotator in a real scenario), that will provide the ground-truth action spots. The set of clips $\mathbf{C}^*$ is manually annotated and both the clips and their corresponding annotations are added to $\mathcal{L}_\tau$.
In a passive learning setup, the oracle would usually randomly select a few clips in $\mathcal{U}_\tau$, potentially annotating redundant information.
In this work, we show that our active learning framework allows us to select relevant clips that increase the performance compared to a random selection, therefore saving time and money.






\section{Experiments}
\label{sec:experiments}


\subsection{Experimental setup}

Our active learning framework is agnostic to the datasets, action spotting training parameters, and active learning selection algorithms. In this section, we provide the technical details describing our experiments in various settings.

\mysection{Datasets.}
In this study, we leverage three datasets to evaluate our active learning framework for action spotting in football videos: SoccerNet-v2, SoccerNet-ball (public), and FWWC19-header (private). Table~\ref{tab:dataset} provides an overview of the main characteristics of each dataset.

\begin{table}[ht]
    \centering
    \resizebox{\columnwidth}{!}{
    \begin{tabular}{l||r|r|r|r}
Dataset          & Games & Annotations & Classes & Density \\ \midrule
SoccerNet-v2     & 550   &   110,458  & 17      &  2.23/min \\ 
SoccerNet-ball   & 9     &    11,041  & 2       & 13.62/min \\ 
FWWC19-header & 52    &     6,527  & 1       &  1.39/min \\ 
    \end{tabular}
    }
    \caption{
    \textbf{Datasets.}
    We investigate our active learning framework on three datasets for action spotting on football videos.
    }
    \label{tab:dataset}
\end{table}

\underline{SoccerNet-v2} consists of $550$ games annotated with $110{,}458$ action spots from $17$ classes of generic actions such as goals, penalties, cards, and free-kicks. These timestamped annotations provide a comprehensive understanding of the actions that occur in football videos.

\underline{SoccerNet-ball} consists of $9$ public games annotated with $11{,}041$ ball-related events such as passes and drives. This dataset provides valuable information on the actions related to the ball, which is a crucial aspect of the game. Moreover, the density of the events in the game requires precise temporal spotting capabilities.

\underline{FWWC19-header} is a private dataset of $52$ games from the FIFA Women World Cup 2019 (FWWC19), annotated for $5$ classes of head impacts, including purposeful headers, unintentional headers, header duels, attempted headers, and other head impacts. This dataset provides insights into the events surrounding head impacts, which are a significant medical concern in football and other contact sports.

\mysection{Action spotting methods.}
In this study, we investigate two action spotting baselines to support our active learning framework for action spotting in football videos: NetVLAD++~\cite{Giancola2021Temporally} and PTS~\cite{Hong2022Spotting-arxiv}. Table~\ref{tab:baseline} provides an overview of the main characteristics of each baseline.

\begin{table}[h]
    \centering
    \resizebox{\columnwidth}{!}{
    \begin{tabular}{l||c|c|c}
Baseline         & Encoder   & AS Head & Training \\ \midrule
NETVLAD++        & ResNet152 & NetVLAD & Head  \\
PTS              & ResNet18  & GRU     & Encoder+Head
    \end{tabular}
    }
    \caption{
    \textbf{Action spotting baselines for football videos.}
    We investigate our active learning framework on two baselines for action spotting on football videos, namely NetVLAD++ and PTS.
    }
    \label{tab:baseline}
\end{table}

\underline{NetVLAD++}~\cite{Giancola2021Temporally} learns to pool temporally contiguous frame features to identify which action class occurs in a clip. Since the feature encoder is frozen and the spotting head is lightweight, it is very fast to train in an active learning fashion. One major feature is that it is trained in a weakly supervised manner that does not take into account the precise localization of the action in the clip, which significantly speeds up the annotation process by the oracle. However, the drawback is that it is less precise in spotting actions.

\underline{Precise Temporal Spotting (PTS)}~\cite{Hong2022Spotting-arxiv} learns to combine dense frame features with a GRU, to identify if specific actions occur on specific frames. The compact encoder is trained end-to-end with the GRU, producing state-of-the-art performances on SoccerNet-v2. 
In this work, we select the ResNet 18 feature encoder that runs on a single GPU, as it is much faster than the RegNet encoder with GRU. 


\mysection{Metrics.}
For action spotting, we rely on the loose Avg-mAP~\cite{Giancola2018SoccerNet}, unless stated otherwise.
For active learning, we analyze the learning curve of the action spotting performance as a function of the ratio of data used to train the model, \ie the size of the labeled dataset. 
Following~\cite{heilbron2018annotate}, we estimate the Area Under the Learning Curve (AULC).
A good active learner is expected to have higher AULC than a random sampler.
Moreover, we propose two more metrics: 
(i) \Md{10}: the Avg-mAP performance when using only 10\% of the data, and
(ii) \Mp{90}: the ratio of data required to reach 90\% of the final Avg-mAP performance.



\mysection{Technical details.}
Unless specified otherwise,
We focus our experiments on the action spotting model NetVLAD++~\cite{Giancola2021Temporally} with the ResNET\_PCA512 features and train the model until convergence using the same training parameters as defined in the original implementation.
At each active learning iteration step, we select an amount of sample $|\mathbf{C^*}|$ equivalent of 1\% of the dataset.
At each action spotting training step, we restart the training from scratch.



\subsection{Initial results}

\begin{figure}[t]
    \centering
    \includegraphics[width=\linewidth,trim={7mm 10mm 0 16mm},clip]{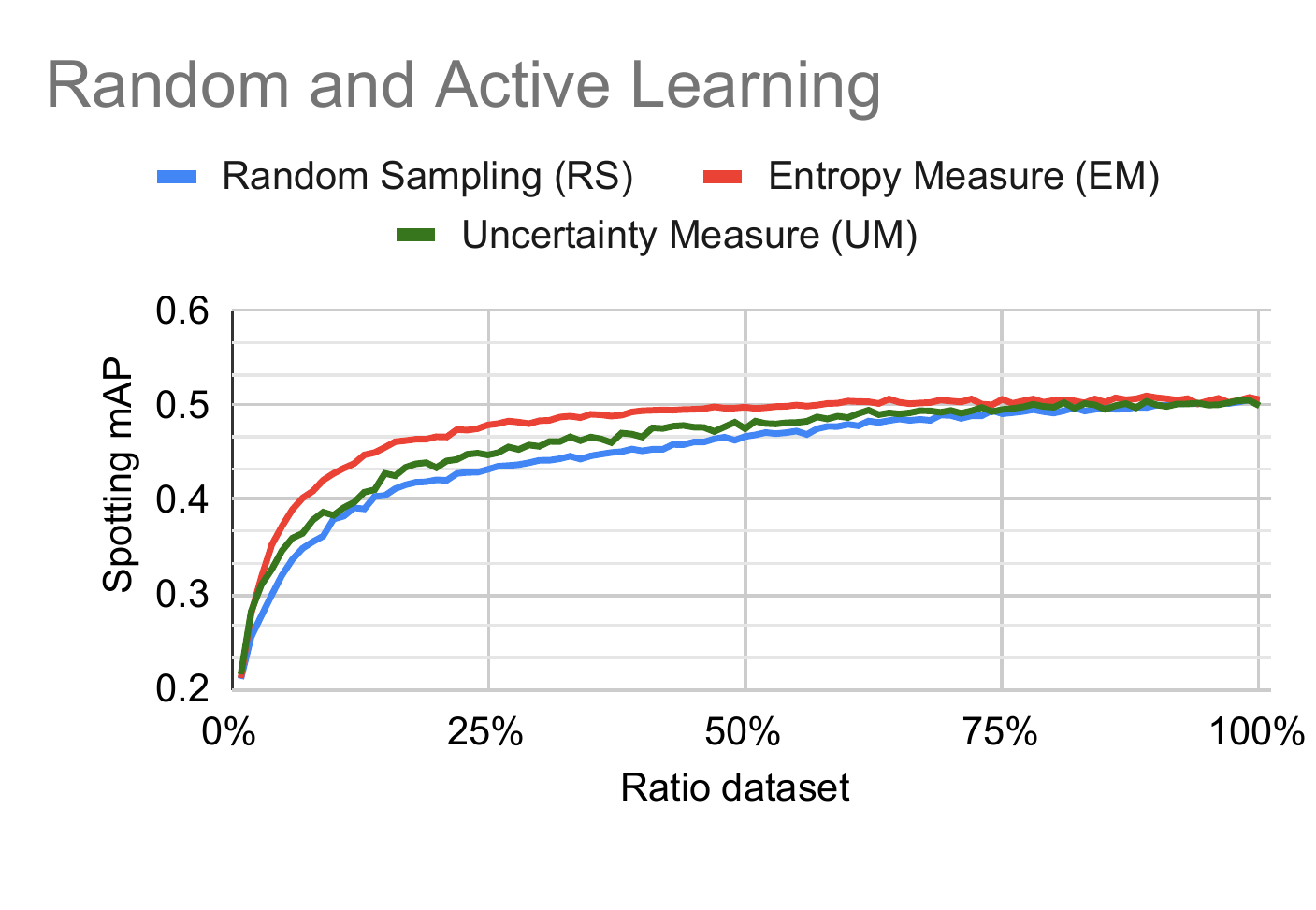}
    \caption{
    \textbf{Active learning \vs random sampling.}
    Our uncertainty sampling using the Entropy Measure (EM) converges to the optimal solution at a faster pace, using fewer data.
    In practice, active learning only needs 36\% of the data needed by a random sampler to reach similar performances (\Mp{90}), and a similar amount of data could lead to up to 18\% performance improvement (\Md{4}).
    }
    \label{fig:MainResults}
\end{figure}

We first compare our framework with two active learning selection algorithms: the Uncertainty Measure (UM) and Entropy Measure (EM), against a random sampler (RS). 
Figure~\ref{fig:MainResults} shows the action spotting performances (loose Avg-mAP) as a function of the labeled dataset size.
Table~\ref{tab:MainResults} reports the main metrics AULC, \Md{} and \Mp{}.
The initial results show that our learning framework significantly accelerates the training. 
With Entropy Measure (EM), the performance of the model converges at a faster pace, thus requiring less annotated data to reach higher performance.
In particular, our setup with EM leads to an AULC metric of 47.96\% \vs 45.11\% with RS.
Moreover, the \Mp{90} of 16\% for EM \vs 45\% for RS indicates that we only need a third of the data to reach 90\% of the action spotting performance.
Finally, the \Md{5} shows that with only 5\% of the data, we reach an action spotting metric Avg-mAP of 37.24\% \vs 32.06\% when sampled randomly.
Interestingly, the Uncertainty Measure (UM) provided only a limited improvement compared to Random Sampling (RS). 
    
\begin{table}[t]
    \centering
    \begin{tabular}{l||c|c|c}
    Method  & RS      & UM      & EM \\
        \midrule
AULC    ($\uparrow$)   & 45.11 & 46.24 & \bf47.96     \\ \midrule
\Mp{90} ($\downarrow$) & 45.00 & 31.00 & \bf16.00 \\ 
\Mp{99} ($\downarrow$) & 99.00 & -     & \bf64.00 \\ \midrule
\Md{5}  ($\uparrow$)   & 32.06 & 34.64 & \bf37.24  \\ 
\Md{10} ($\uparrow$)   & 37.98 & 38.38 & \bf42.79  \\
\end{tabular}
    \caption{
    \textbf{Active learning \vs random sampling.}
    Our proposed active learning framework based on Entropy Measure (EM) outperforms Random Sampling (RS) and Uncertainty Measure (UM).
    }
    \label{tab:MainResults}
\end{table}

\subsection{Accelerating the active learning framework}

In this section, we share a few findings that speed up our active learning framework. In particular, 
\textbf{(i)} we introduce a faster scheduler for NetVLAD++ that lead to similar performance,
\textbf{(ii)} we introduce an Adaptive Active Learning scheduler (AdapAL),
\textbf{(iii)} we investigate a continual training that fine-tunes the model instead of training from scratch at each active learning step.

\begin{figure}[t]
    \centering
    \includegraphics[width=\linewidth,trim={7mm 10mm 0 16mm},clip]{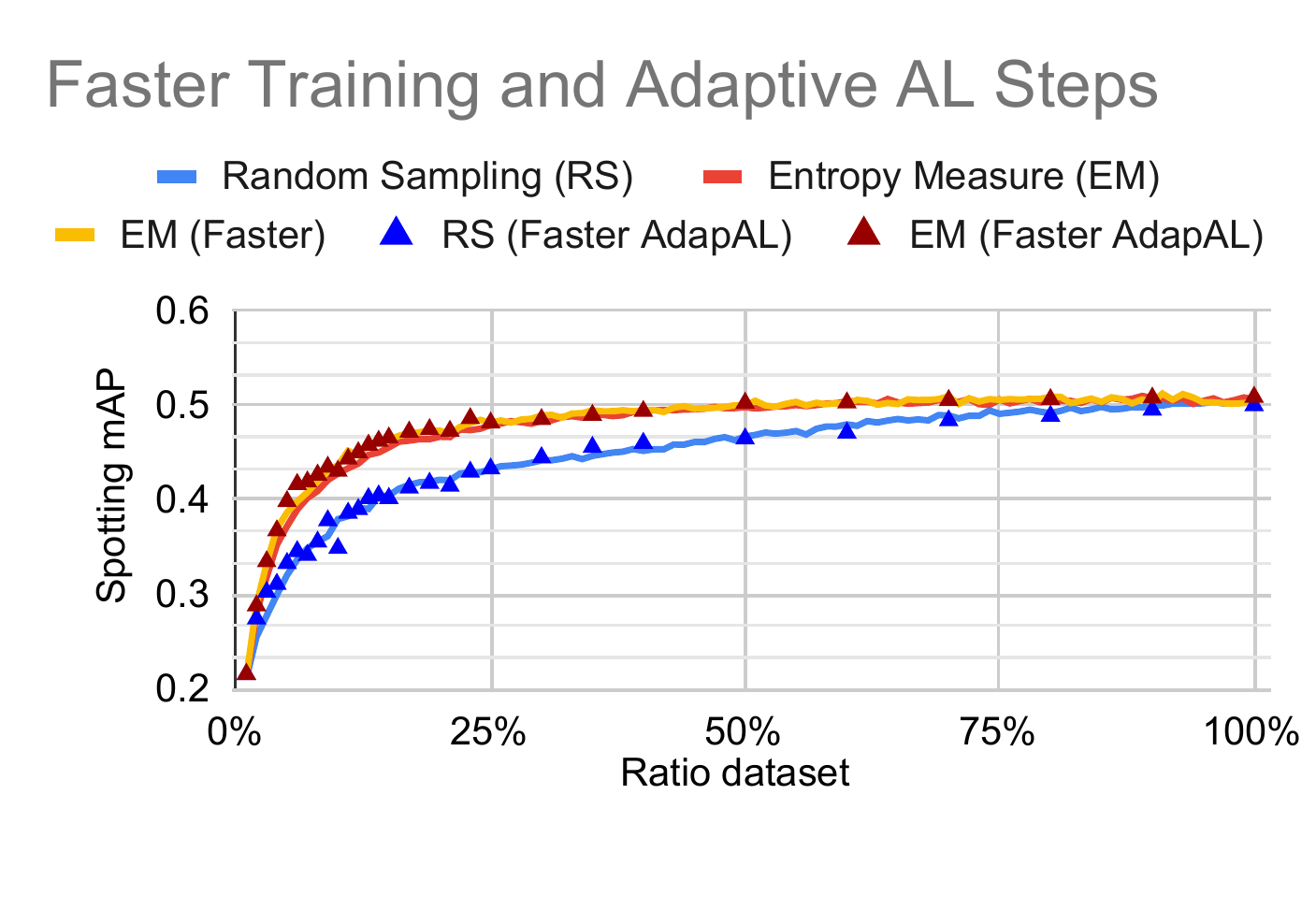}
    \caption{
    \textbf{Faster training and adaptive active learning (AdapAL) paradigms.}
    We show here that we can significantly decrease the active learning time for our experiments without reducing in any way the performance of the active learning training.
    }
    \label{fig:FasterTraining}
\end{figure}

\mysection{Faster model training.}
First, we leverage a faster scheduler for the learning rate. Instead of starting from $10^{-3}$, and reducing the learning rate on each validation loss plateau with a ratio of $10$, until we reach $10^{-8}$, we start with a learning rate of $10^{-2}$ and reduce it down to $10^{-4}$. Also, we reduce the patience to identify the plateau from $10$ to $5$ epochs. 
By doing so, we practically cut the training time by two, still producing performance on par with the original training scheduler, as shown in Figure~\ref{fig:FasterTraining}.

\mysection{Adaptive active learning scheduler.}
Second, we adapt the active learning (AL) steps, gradually increasing the number of samples selected and annotated per AL step. 
At regime, we can increase the dataset $\mathcal{L}_t$ by more than only 1\% of the dataset. 
In practice, we chose to increment 2\% after 15\% of the dataset, 5\% after 25\% of the dataset, and 10\% after 40\% of the dataset.
By doing so, we reduced the AL steps from $100$ to $30$, saving 70\% of the time.
Figure~\ref{fig:FasterTraining} illustrates that the Adaptive AL step size does not impact the performance of the training of NetVLAD++ on SoccerNet-v2.

\begin{figure}[t]
    \centering
    \includegraphics[width=\linewidth,trim={7mm 10mm 0 16mm},clip]{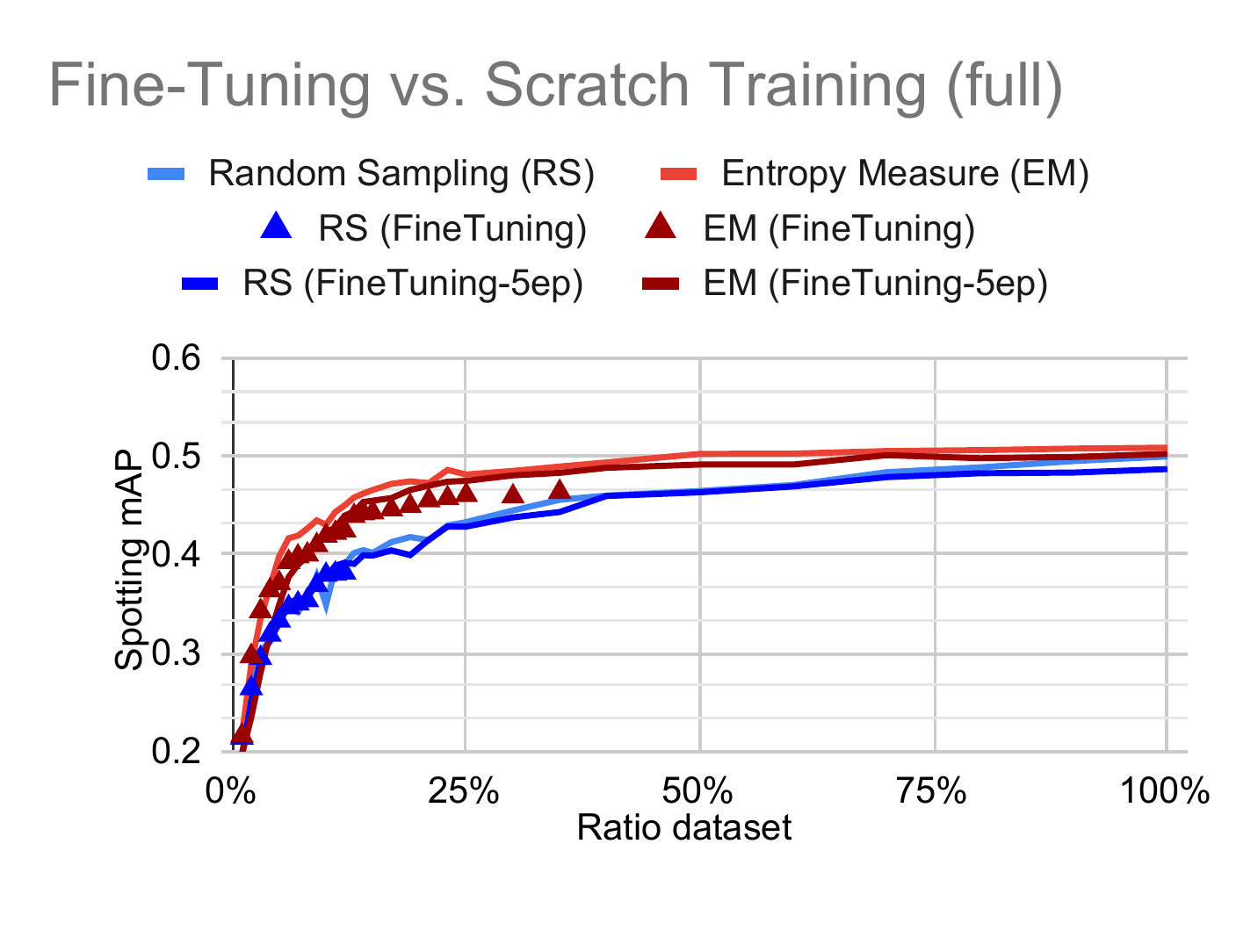}
    \caption{
    \textbf{Effect of fine-tuning a limited number of epochs.}
    Fine-tuning from a model with a limited number of epochs leads to more stability in the training for the next active learning step. 
    }
    \label{fig:FineTuning}
\end{figure}

\begin{table}[t]
    \centering
    \resizebox{\columnwidth}{!}{
    \begin{tabular}{l|l||c|c|c}
AL Setup  & Train &  AULC ($\uparrow$) & \Md{10} ($\uparrow$) & \Mp{90} ($\downarrow$) \\ \midrule
RS$_{1\%   }$ & orig.    &   45.11 &   37.98 &   45.00  \\ 
EM$_{1\%   }$ & orig.    &   47.96 &   42.79 &   16.00  \\
EM$_{1\%   }$ & fast     &\bf48.28 &\bf43.68 &\bf14.00  \\ \midrule
RS$_{AdapAL}$ & fast     &   44.64 &   34.91 &   40.00  \\ 
EM$_{AdapAL}$ & fast     &\bf48.01 &\bf43.06 &\bf13.00  \\ \midrule
RS$_{AdapAL}$ & $5$ ep.  &   44.13 &   37.78 &   40.00  \\ 
EM$_{AdapAL}$ & $5$ ep.  &\bf46.94 &\bf42.30 &\bf19.00  \\ 
    \end{tabular}
    }
    \caption{
    \textbf{Ablation.}
    Our proposed active learning framework based on Entropy Measure (EM) outperforms Random Sampling (RS) on all active learning setups.
    }
    \label{tab:Ablation}
\end{table}

\mysection{Continual training.}
Third, we investigate whether resuming the training from the previous active learning step would be beneficial to reduce the training time.
The stopping series of triangles in Figure~\ref{fig:FineTuning} illustrates that a naive implementation of continual training with the original parameters leads to a divergent loss that impedes any further fine-tuning.
Instead, we propose to bootstrap the training with 20 epochs and fine-tune the model for 5 epochs at each active learning step, with a LR fixed at $10^{-2}$.
We can see that continuing the training for 5 epoch in every active learning step still preserve the same trend of EM outperforming RS.

Table~\ref{tab:Ablation} summarizes how (i) training faster, (ii) adapting the active learning scheduler, and (iii) continuing the training actually performs in terms of metrics. Despite the minimal difference in the performances, the gap between RS and EM is maintained. Most importantly, those tricks lead to an order of magnitude acceleration in running the experiments.

\subsection{Generalization analyses}

\begin{table}[t]
    \centering
    \resizebox{\columnwidth}{!}{
    \begin{tabular}{l|c||c|c|c}
Data              & Metric                 & RS      &   EM    &  UM     \\ 
\midrule
SoccerNet-ball    & AULC ($\uparrow$)      &   40.47 &   42.18 &\bf42.88  \\ 
SoccerNet-ball    & \Md{10} ($\uparrow$)   &   36.55 &\bf42.41 &   41.88  \\ 
SoccerNet-ball    & \Mp{90} ($\downarrow$) &   23.00 &\bf 8.00 &    9.00  \\ 
\midrule
FWWC19-header       & AULC ($\uparrow$)      &   42.67 &   44.28 &\bf44.59  \\ 
FWWC19-header       & \Md{10} ($\uparrow$)   &   35.18 &   42.32 &\bf42.39  \\  
FWWC19-header       & \Mp{90} ($\downarrow$) &   35.00 &\bf12.00 &\bf12.00  \\  
    \end{tabular}
    }
    \caption{
    \textbf{Dataset generalization.}
    EM and UM outperform RS on the other two datasets.
    With less class diversity, the gap between EM and UM is smaller.
    }
    \label{tab:MoreDatasets}
\end{table}

\mysection{Datasets generalization.}
We successively experimented our framework on two other datasets, namely SoccerNet-ball and FWWC19-header.
Since SoccerNet-ball has denser actions, the hyper-parameters of NetVLAD++ were refined with a temporal window of 1s and an NMS of 1s.
We chose the accelerated active learning settings, with a faster training scheduler, adaptive active learning step, and continual fine-tuning for 5 epochs per active learning step, after a bootstrap of 20 epochs.
Table~\ref{tab:MoreDatasets} details the results and shows, in particular, that UM and EM significantly accelerate the training efficiency for both  datasets compared to Random Sampling.
Interestingly, the gap between UM and EM is smaller in these two datasets than it was on SoccerNet-v2.
We hypothesize that this behavior originates from the lower number of classes in SoccerNet-Headers and FWWC19-Header, respectively 2 and 1 (see Table~\ref{tab:dataset}).
In fact, the ranking for the samples from UM and EM are actually similar in the case of a binary classifier (see Equations~\eqref{eq:UM} and \eqref{eq:EM}).


\mysection{Architecture generalization.}
We analyzed the generalization capability of our active learning framework to other action spotting methods, in particular PTS~\cite{Hong2022Spotting-arxiv}.
Unlike NetVLAD++, PTS produces class prediction scores per frame instead of per clip.
To estimate an active learning score per clip, we aggregate the active learning score per frame with mean or max pooling.
The former will consider an average uncertainty along all frames of the clip, the latter will sample clips containing single uncertain frames to train next.
Similarly, we chose the same accelerated active learning settings, as PTS is way slower to train than NetVLAD++.
Aligned with the findings from NetVLAD++, we show in Table~\ref{tab:Ablation} that UM and EM outperform RS on all three datasets.
Moreover, we identify a similar trend showing that with a lower number of classes, UM tends to outperform EM.
Finally, max pooling appears to work better, which means that clips with single uncertain frames are generally more informative to train the model.

\begin{table}[t]
    \centering
    \resizebox{\columnwidth}{!}{
    \begin{tabular}{l|c||c|c|c}
Dataset & AL  & AULC ($\uparrow$) & \Md{10} ($\uparrow$) & \Mp{90} ($\downarrow$) \\ \midrule
SoccerNet-v2     & RS      &   27.76 &   13.59  &   60.00 \\ 
SoccerNet-v2     & mean-EM &   28.53 &   16.14  &\bf50.00 \\ 
SoccerNet-v2     & ~max-EM &\bf28.83 &\bf17.62  &\bf50.00 \\ 
SoccerNet-v2     & mean-UM &   28.26 &   13.58  &   60.00     \\ 
SoccerNet-v2     & ~max-UM &   28.80 &   15.64  &\bf50.00 \\ \midrule
SoccerNet-ball   & RS      &   56.37 &   52.28  &   19.00 \\ 
SoccerNet-ball   & mean-EM &   54.26 &   46.57  &   40.00 \\ 
SoccerNet-ball   & ~max-EM &   57.64 &   51.79  &   25.00 \\ 
SoccerNet-ball   & mean-UM &   54.92 &   49.53  &   35.00 \\ 
SoccerNet-ball   & ~max-UM &\bf58.48 &\bf53.80  &\bf12.00  \\ \midrule
FWWC19-header      & RS      &   29.77 &   21.77  &   -     \\ 
FWWC19-header      & mean-EM &   34.81 &   18.45  &\bf40.00 \\ 
FWWC19-header      & ~max-EM &   33.92 &   23.43  &   70.00 \\ 
FWWC19-header      & mean-UM &\bf35.93 &   19.14  &   50.00 \\ 
FWWC19-header      & ~max-UM &   35.26 &\bf24.86  &\bf40.00 \\
    \end{tabular} 
    }
    \caption{
    \textbf{Architecture generalization.} 
    The EM and UM active selection function also outperform the RS selection when coupled with PTS~\cite{Hong2022Spotting-arxiv}.
    }
    \label{tab:ArchitectureGeneralization}
\end{table}

\section{Conclusion}
\label{sec:conclusion}

In conclusion, our proposed active learning framework selects the most informative video samples to be annotated next, thus reducing the annotation effort and accelerating the training of action spotting models. We leveraged uncertainty sampling to select the most challenging video clip to train on next, which speeds up the learning process of the models. We show that our framework effectively reduces the required training data for accurate action spotting in football videos, achieving similar performance with NetVLAD++ on SoccerNet-v2 using only one-third of the dataset. This indicates significant capabilities for reducing annotation effort. Furthermore, we validated our approach on two new datasets that focus on localizing in time the actions of headers and passes. 
In future works, we will investigate the use of other active learning paradigms for the task of action spotting, such as diversity maximization, query-by-committee, and expected error. 

{

\mysection{Acknowledgement.}
This work was partly supported by the King Abdullah University of Science and Technology (KAUST) Office of Sponsored Research through the Visual Computing Center (VCC) funding and the SDAIA-KAUST Center of Excellence in Data Science and Artificial Intelligence (SDAIA-KAUST AI). 
A. Cioppa is funded by the F.R.S.-FNRS.
We thank Eloise Arnold, who helped design the reliability protocol for the FWWC19-header dataset.
}



\clearpage
{\small
\bibliographystyle{ieee_fullname}
\bibliography{bib/abbreviation-short,bib/active,bib/bib,bib/dataset,bib/egbib,bib/labo,bib/soccer,bib/sports}

\begin{thebibliography}{10}\itemsep=-1pt

\bibitem{ArbuesSanguesa2020Using}
Adri{\`a} Arbu{\'e}s~Sang{\"u}esa, Adri{\`a}n Mart{\'i}n, Javier Fern{\'a}ndez,
  Coloma Ballester, and Gloria Haro.
\newblock Using player's body-orientation to model pass feasibility in soccer.
\newblock In {\em IEEE/CVF Conf. Comput. Vis. Pattern Recognit. Work. (CVPRW)},
  pages 3875--3884, Seattle, WA, USA, Jun. 2020. Inst. Electr. Electron. Eng.
  (IEEE).

\bibitem{bandla2013active}
Sunil Bandla and Kristen Grauman.
\newblock Active learning of an action detector from untrimmed videos.
\newblock In {\em IEEE Int. Conf. Comput. Vis. (ICCV)}, pages 1833--1840,
  Sydney, NSW, Australia, Dec. 2013. Inst. Electr. Electron. Eng. (IEEE).

\bibitem{Caba2015ActivityNet}
Fabian Caba~Heilbron, Victor Escorcia, Bernard Ghanem, and Juan Carlos~Niebles.
\newblock {ActivityNet}: A large-scale video benchmark for human activity
  understanding.
\newblock In {\em IEEE/CVF Conf. Comput. Vis. Pattern Recognit. (CVPR)}, pages
  961--970, Boston, MA, USA, Jun. 2015. Inst. Electr. Electron. Eng. (IEEE).

\bibitem{Cartas2022AGraphbased}
Alejandro Cartas, Coloma Ballester, and Gloria Haro.
\newblock A graph-based method for soccer action spotting using unsupervised
  player classification.
\newblock In {\em Int. ACM Work. Multimedia Content Anal. Sports (MMSports)},
  pages 93--102, Lisbon, Port., Oct. 2022. ACM.

\bibitem{Cho2014OnThe}
Kyunghyun Cho, Bart van Merrienboer, Dzmitry Bahdanau, and Yoshua Bengio.
\newblock On the properties of neural machine translation: Encoder-decoder
  approaches.
\newblock In {\em Proc. SSST-8, Eighth Work. Syntax. Semant. Struct. Stat.
  Transl.}, pages 103--111, Doha, Qatar, 2014. Association for Computational
  Linguistics.

\bibitem{Cioppa2022Scaling}
Anthony Cioppa, Adrien Deli{\`e}ge, Silvio Giancola, Bernard Ghanem, and Marc
  Van~Droogenbroeck.
\newblock Scaling up {SoccerNet} with multi-view spatial localization and
  re-identification.
\newblock {\em Sci. Data}, 9(1):1--9, Jun. 2022.

\bibitem{Cioppa2020AContextAware}
Anthony Cioppa, Adrien Deli\`ege, Silvio Giancola, Bernard Ghanem, Marc
  Van~Droogenbroeck, Rikke Gade, and Thomas~B. Moeslund.
\newblock A context-aware loss function for action spotting in soccer videos.
\newblock In {\em IEEE/CVF Conf. Comput. Vis. Pattern Recognit. (CVPR)}, pages
  13123--13133, Seattle, WA, USA, Jun. 2020. Inst. Electr. Electron. Eng.
  (IEEE).

\bibitem{Cioppa2022SoccerNetTracking}
Anthony Cioppa, Silvio Giancola, Adrien Deliege, Le Kang, Xin Zhou, Zhiyu
  Cheng, Bernard Ghanem, and Marc Van~Droogenbroeck.
\newblock {SoccerNet}-tracking: Multiple object tracking dataset and benchmark
  in soccer videos.
\newblock In {\em IEEE Int. Conf. Comput. Vis. Pattern Recognit. Work. (CVPRW),
  CVsports}, pages 3490--3501, New Orleans, LA, USA, Jun. 2022. Inst. Electr.
  Electron. Eng. (IEEE).

\bibitem{Darwish2022STE}
Abdulrahman Darwish and Tallal El-Shabrway.
\newblock {STE}: Spatio-temporal encoder for action spotting in soccer videos.
\newblock In {\em Int. ACM Work. Multimedia Content Anal. Sports (MMSports)},
  pages 87--92, Lisbon, Port., Oct. 2022. ACM.

\bibitem{Decroos2019Actions}
Tom Decroos, Lotte Bransen, Jan Van~Haaren, and Jesse Davis.
\newblock Actions speak louder than goals.
\newblock In {\em ACM SIGKDD Int. Conf. Knowl. Discov. \& Data Min.}, page
  1851–1861. ACM, Jul. 2019.

\bibitem{Deliege2021SoccerNetv2}
Adrien Deli{\`e}ge, Anthony Cioppa, Silvio Giancola, Meisam~J. Seikavandi,
  Jacob~V. Dueholm, Kamal Nasrollahi, Bernard Ghanem, Thomas~B. Moeslund, and
  Marc Van~Droogenbroeck.
\newblock {SoccerNet}-v2: A dataset and benchmarks for holistic understanding
  of broadcast soccer videos.
\newblock In {\em IEEE Int. Conf. Comput. Vis. Pattern Recognit. Work. (CVPRW),
  CVsports}, pages 4508--4519, Nashville, TN, USA, Jun. 2021.
\newblock Best CVSports paper award.

\bibitem{duros2017human}
Vasilios Duros, Jonathan Grizou, Weimin Xuan, Zied Hosni, De-Liang Long,
  Haralampos~N Miras, and Leroy Cronin.
\newblock Human versus robots in the discovery and crystallization of gigantic
  polyoxometalates.
\newblock {\em Angewandte Chemie}, 129(36):10955--10960, 2017.

\bibitem{Ekin2003Automatic}
A. Ekin, A.M. Tekalp, and R. Mehrotra.
\newblock Automatic soccer video analysis and summarization.
\newblock {\em IEEE Trans. Image Process.}, 12(7):796--807, Jul. 2003.

\bibitem{freund1997selective}
Yoav Freund, Sebastian~H. Seung, Eli Shamir, and Naftali Tishby.
\newblock Selective sampling using the query by committee algorithm.
\newblock {\em Machine learning}, 28(2-3):133--168, Aug. 1997.

\bibitem{Castanon2016Exploratory}
Gregory G.~Castanon, Casta\~n\'on.
\newblock {\em Exploratory search through large video corpora}.
\newblock PhD thesis, Boston University, USA, 2016.

\bibitem{gal2017deep}
Yarin Gal, Riashat Islam, and Zoubin Ghahramani.
\newblock Deep {Bayesian} active learning with image data.
\newblock In {\em Int. Conf. Mach. Learn. (ICML)}, pages 1183--1192. PMLR,
  2017.

\bibitem{Giancola2018SoccerNet}
Silvio Giancola, Mohieddine Amine, Tarek Dghaily, and Bernard Ghanem.
\newblock {SoccerNet}: A scalable dataset for action spotting in soccer videos.
\newblock In {\em IEEE/CVF Conf. Comput. Vis. Pattern Recognit. Work. (CVPRW)},
  pages 1792--179210, Salt Lake City, UT, USA, Jun. 2018. Inst. Electr.
  Electron. Eng. (IEEE).

\bibitem{Giancola2022SoccerNet}
Silvio Giancola, Anthony Cioppa, Adrien Deli{\`e}ge, Floriane Magera, Vladimir
  Somers, Le Kang, Xin Zhou, Olivier Barnich, Christophe De~Vleeschouwer,
  Alexandre Alahi, Bernard Ghanem, Marc Van~Droogenbroeck, Abdulrahman Darwish,
  Adrien Maglo, Albert Clap{\'e}s, Andreas Luyts, Andrei Boiarov, Artur Xarles,
  Astrid Orcesi, Avijit Shah, Baoyu Fan, Bharath Comandur, Chen Chen, Chen
  Zhang, Chen Zhao, Chengzhi Lin, Cheuk-Yiu Chan, Chun~Chuen Hui, Dengjie Li,
  Fan Yang, Fan Liang, Fang Da, Feng Yan, Fufu Yu, Guanshuo Wang, H.~Anthony
  Chan, He Zhu, Hongwei Kan, Jiaming Chu, Jianming Hu, Jianyang Gu, Jin Chen,
  Jo{\~a}o V.~B. Soares, Jonas Theiner, Jorge De~Corte, Jos{\'e}~Henrique
  Brito, Jun Zhang, Junjie Li, Junwei Liang, Leqi Shen, Lin Ma, Lingchi Chen,
  Miguel Santos~Marques, Mike Azatov, Nikita Kasatkin, Ning Wang, Qiong Jia,
  Quoc~Cuong Pham, Ralph Ewerth, Ran Song, Rengang Li, Rikke Gade, Ruben
  Debien, Runze Zhang, Sangrok Lee, Sergio Escalera, Shan Jiang, Shigeyuki
  Odashima, Shimin Chen, Shoichi Masui, Shouhong Ding, Sin-wai Chan, Siyu Chen,
  Tallal El-Shabrawy, Tao He, Thomas~B. Moeslund, Wan-Chi Siu, Wei Zhang, Wei
  Li, Xiangwei Wang, Xiao Tan, Xiaochuan Li, Xiaolin Wei, Xiaoqing Ye, Xing
  Liu, Xinying Wang, Yandong Guo, Yaqian Zhao, Yi Yu, Yingying Li, Yue He,
  Yujie Zhong, Zhenhua Guo, and Zhiheng Li.
\newblock {SoccerNet} 2022 challenges results.
\newblock In {\em Int. ACM Work. Multimedia Content Anal. Sports (MMSports)},
  pages 75--86, Lisbon, Port., Oct. 2022. ACM.

\bibitem{Giancola2021Temporally}
Silvio Giancola and Bernard Ghanem.
\newblock Temporally-aware feature pooling for action spotting in soccer
  broadcasts.
\newblock In {\em IEEE Int. Conf. Comput. Vis. Pattern Recognit. (CVPR)}, pages
  4490--4499, Nashville, TN, USA, Jun. 2021.

\bibitem{gilad2005query}
Ran Gilad-Bachrach, Amir Navot, and Naftali Tishby.
\newblock Query by committee made real.
\newblock In {\em NeurIPS}, volume~18, 2005.

\bibitem{hakkani2002active}
Dilek Hakkani-T{\"u}r, Giuseppe Riccardi, and Allen Gorin.
\newblock Active learning for automatic speech recognition.
\newblock In {\em IEEE Int. Conf. Acoust. Speech Signal Process. (ICASSP)},
  volume~4. IEEE, 2002.

\bibitem{heilbron2018annotate}
Fabian~Caba Heilbron, Joon-Young Lee, Hailin Jin, and Bernard Ghanem.
\newblock What do {I} annotate next? an empirical study of active learning for
  action localization.
\newblock In {\em ECCV}, volume 11215 of {\em Lect. Notes Comput. Sci.}, pages
  212--229. Springer Int. Publ., 2018.

\bibitem{hoi2006batch}
Steven~C. Hoi, Rong Jin, Jianke Zhu, and Michael~R. Lyu.
\newblock Batch mode active learning and its application to medical image
  classification.
\newblock In {\em Int. Conf. Mach. Learn. (ICML)}, pages 417--424, 2006.

\bibitem{Hong2022Spotting-arxiv}
James Hong, Haotian Zhang, Micha{\"e}l Gharbi, Matthew Fisher, and Kayvon
  Fatahalian.
\newblock Spotting temporally precise, fine-grained events in video.
\newblock {\em CoRR}, abs/2207.10213, 2022.

\bibitem{Host2022AnOverview}
Kristina Host and Marina Iva{\v s}i{\' c}-Kos.
\newblock An overview of human action recognition in sports based on computer
  vision.
\newblock {\em Heliyon}, 8(6), Jun. 2022.

\bibitem{houlsby2011bayesian}
Neil Houlsby, Ferenc Husz{\'a}r, Zoubin Ghahramani, and M{\'a}t{\'e} Lengyel.
\newblock Bayesian active learning for classification and preference learning.
\newblock {\em CoRR}, abs/1112.5745, 2011.

\bibitem{IFAB2022Laws}
IFAB.
\newblock Laws of the game.
\newblock Technical report, The International Football Association Board,
  Zurich, Switzerland, 2022.

\bibitem{iglesias2011combining}
Juan~Eugenio Iglesias, Ender Konukoglu, Albert Montillo, Zhuowen Tu, and
  Antonio Criminisi.
\newblock Combining generative and discriminative models for semantic
  segmentation of {CT} scans via active learning.
\newblock In {\em Information Processing in Medical Imaging}, volume 6801 of
  {\em Lect. Notes Comput. Sci.}, pages 25--36. Springer Berlin Heidelberg,
  2011.

\bibitem{Jiang2020SoccerDB}
Yudong Jiang, Kaixu Cui, Leilei Chen, Canjin Wang, and Changliang Xu.
\newblock {SoccerDB}: A large-scale database for comprehensive video
  understanding.
\newblock In {\em Int. ACM Work. Multimedia Content Anal. Sports (MMSports)},
  page 1–8. ACM, Oct. 2020.

\bibitem{joshi2009multi}
Ajay~J. Joshi, Fatih Porikli, and Nikolaos Papanikolopoulos.
\newblock Multi-class active learning for image classification.
\newblock In {\em IEEE/CVF Conf. Comput. Vis. Pattern Recognit. (CVPR)}, pages
  2372--2379, Miami, FL, USA, Jun. 2009. Inst. Electr. Electron. Eng. (IEEE).

\bibitem{joshi2012scalable}
Ajay~J. Joshi, Fatih Porikli, and Nikolaos~P. Papanikolopoulos.
\newblock Scalable active learning for multiclass image classification.
\newblock {\em IEEE Trans. Pattern Anal. Mach. Intell.}, 34(11):2259--2273,
  Nov. 2012.

\bibitem{konyushkova2017learning}
Ksenia Konyushkova, Raphael Sznitman, and Pascal Fua.
\newblock Learning active learning from data.
\newblock In {\em NeurIPS}, volume~30, 2017.

\bibitem{lewis1994heterogeneous}
David~D. Lewis and Jason Catlett.
\newblock Heterogeneous uncertainty sampling for supervised learning.
\newblock {\em Mach. Learn. Proc.}, pages 148--156, 1994.

\bibitem{Liu2022MonoTrack}
Paul Liu and Jui-Hsien Wang.
\newblock {MonoTrack}: Shuttle trajectory reconstruction from monocular
  badminton video.
\newblock In {\em IEEE/CVF Conf. Comput. Vis. Pattern Recognit. Work. (CVPRW)},
  pages 3512--3521, New Orleans, LA, USA, Jun. 2022. Inst. Electr. Electron.
  Eng. (IEEE).

\bibitem{Maglo2022Efficient}
Adrien Maglo, Astrid Orcesi, and Quoc-Cuong Pham.
\newblock Efficient tracking of team sport players with few game-specific
  annotations.
\newblock In {\em IEEE/CVF Conf. Comput. Vis. Pattern Recognit. Work. (CVPRW)},
  pages 3460--3470, New Orleans, LA, USA, Jun. 2022. Inst. Electr. Electron.
  Eng. (IEEE).

\bibitem{Moeslund2014Computer}
Thomas~B. Moeslund, Graham Thomas, and Adrian Hilton.
\newblock {\em Computer vision in sports}.
\newblock Springer, 2014.

\bibitem{nonaka2022end}
Naoki Nonaka, Ryo Fujihira, Monami Nishio, Hidetaka Murakami, Takuya Tajima,
  Mutsuo Yamada, Akira Maeda, and Jun Seita.
\newblock End-to-end high-risk tackle detection system for rugby.
\newblock In {\em Proceedings of the IEEE/CVF Conference on Computer Vision and
  Pattern Recognition}, pages 3550--3559, 2022.

\bibitem{Pappalardo2019APublic}
Luca Pappalardo, Paolo Cintia, Alessio Rossi, Emanuele Massucco, Paolo
  Ferragina, Dino Pedreschi, and Fosca Giannotti.
\newblock A public data set of spatio-temporal match events in soccer
  competitions.
\newblock {\em Sci. Data}, 6(1):1--15, Oct. 2019.

\bibitem{Pavel2022VisionBased}
Monirul~Islam Pavel, Siok~Yee Tan, and Azizi Abdullah.
\newblock Vision-based autonomous vehicle systems based on deep learning: A
  systematic literature review.
\newblock {\em Appl. Sci.}, 12(14):1--51, Jul. 2022.

\bibitem{Ren2021ASurvey}
Pengzhen Ren, Yun Xiao, Xiaojun Chang, Po-Yao Huang, Zhihui Li, Brij~B. Gupta,
  Xiaojiang Chen, and Xin Wang.
\newblock A survey of deep active learning.
\newblock {\em ACM Comput. Surv.}, 54(9):1--40, Oct. 2021.

\bibitem{Rongved2021Automated}
Olav Rongved, Markus Stige, Steven Hicks, Vajira Thambawita, Cise Midoglu, Evi
  Zouganeli, Dag Johansen, Michael Riegler, and P{\r{a}}l Halvorsen.
\newblock Automated event detection and classification in soccer: The potential
  of using multiple modalities.
\newblock {\em Machine Learning and Knowledge Extraction}, 3(4):1--25, Dec.
  2021.

\bibitem{Scott2022SoccerTrack}
Atom Scott, Ikuma Uchida, Masaki Onishi, Yoshinari Kameda, Kazuhiro Fukui, and
  Keisuke Fujii.
\newblock {SoccerTrack}: A dataset and tracking algorithm for soccer with
  fish-eye and drone videos.
\newblock In {\em IEEE/CVF Conf. Comput. Vis. Pattern Recognit. Work. (CVPRW)},
  pages 3568--3578, New Orleans, LA, USA, Jun. 2022. Inst. Electr. Electron.
  Eng. (IEEE).

\bibitem{sener2017active}
Ozan Sener and Silvio Savarese.
\newblock Active learning for convolutional neural networks: A core-set
  approach.
\newblock {\em CoRR}, abs/1708.00489, 2017.

\bibitem{settles2009active}
Burr Settles.
\newblock Active learning literature survey.
\newblock Computer Sciences Technical Report 1648, 2009.

\bibitem{seung1992query}
H.~Sebastian Seung, Manfred Opper, and Haim Sompolinsky.
\newblock Query by committee.
\newblock In {\em Proc. Fifth Annu. Work. Comput. Learn. Theory}, page
  287–294. ACM, Jul. 1992.

\bibitem{Soares2022Action-arxiv}
Jo{\~a}o V.~B. Soares and Avijit Shah.
\newblock Action spotting using dense detection anchors revisited: Submission
  to the {SoccerNet} challenge 2022.
\newblock {\em CoRR}, abs/2206.07846, 2022.

\bibitem{Soares2022Temporally}
Joao V.~B. Soares, Avijit Shah, and Topojoy Biswas.
\newblock Temporally precise action spotting in soccer videos using dense
  detection anchors.
\newblock In {\em IEEE Int. Conf. Image Process. (ICIP)}, pages 2796--2800,
  Bordeaux, France, Oct. 2022. Inst. Electr. Electron. Eng. (IEEE).

\bibitem{Soldan2021MADAS}
Mattia Soldan, A. Pardo, Juan~Le'on Alc'azar, Fabian~Caba Heilbron, Chen Zhao,
  Silvio Giancola, and Bernard Ghanem.
\newblock Mad: A scalable dataset for language grounding in videos from movie
  audio descriptions.
\newblock {\em 2022 IEEE/CVF Conference on Computer Vision and Pattern
  Recognition (CVPR)}, pages 5016--5025, 2021.

\bibitem{Somers2023Body}
Vladimir Somers, Christophe De~Vleeschouwer, and Alexandre Alahi.
\newblock Body part-based representation learning for occluded person
  {Re-Identification}.
\newblock In {\em IEEE/CVF Winter Conf. Appl. Comput. Vis. (WACV)}, pages
  1613--1623, Waikoloa, HI, USA, Jan. 2023. Inst. Electr. Electron. Eng.
  (IEEE).

\bibitem{Sreenu2019Intelligent}
G. Sreenu and M.~A. Saleem~Durai.
\newblock Intelligent video surveillance: a review through deep learning
  techniques for crowd analysis.
\newblock {\em J. Big Data}, 6(1), Jun. 2019.

\bibitem{Sudhakaran2020GateShift}
Swathikiran Sudhakaran, Sergio Escalera, and Oswald Lanz.
\newblock Gate-shift networks for video action recognition.
\newblock In {\em IEEE/CVF Conf. Comput. Vis. Pattern Recognit. (CVPR)}, pages
  1099--1108, Seattle, WA, USA, Jun. 2020. Inst. Electr. Electron. Eng. (IEEE).

\bibitem{Suzuki2019Team}
Genki Suzuki, Sho Takahashi, Takahiro Ogawa, and Miki Haseyama.
\newblock Team tactics estimation in soccer videos based on a deep extreme
  learning machine and characteristics of the tactics.
\newblock {\em IEEE Access}, 7:153238--153248, 2019.

\bibitem{Thomas2017Computer}
Graham Thomas, Rikke Gade, Thomas~B. Moeslund, Peter Carr, and Adrian Hilton.
\newblock Computer vision for sports: current applications and research topics.
\newblock {\em Comput. Vis. Image Underst.}, 159:3--18, Jun. 2017.

\bibitem{thompson1999active}
Cynthia~A. Thompson, Mary~Elaine Califf, and Raymond~J. Mooney.
\newblock Active learning for natural language parsing and information
  extraction.
\newblock In {\em Int. Conf. Mach. Learn. (ICML)}, pages 406--414. Citeseer,
  1999.

\bibitem{Tomei2021RMSNet}
Matteo Tomei, Lorenzo Baraldi, Simone Calderara, Simone Bronzin, and Rita
  Cucchiara.
\newblock {RMS}-net: Regression and masking for soccer event spotting.
\newblock In {\em IEEE Int. Conf. Pattern Recognit. (ICPR)}, pages 7699--7706,
  Milan, Italy, Jan. 2021. Inst. Electr. Electron. Eng. (IEEE).

\bibitem{tong2001support}
Simon Tong and Edward Chang.
\newblock Support vector machine active learning for image retrieval.
\newblock In {\em Proc. Ninth ACM Int. Conf. Multimedia}, page 107–118. ACM,
  Oct. 2001.

\bibitem{VanZandycke2022DeepSportradarv1}
Gabriel Van~Zandycke, Vladimir Somers, Maxime Istasse, Carlo~Del Don, and
  Davide Zambrano.
\newblock {DeepSportradar}-v1: Computer vision dataset for sports understanding
  with high quality annotations.
\newblock In {\em Int. ACM Work. Multimedia Content Anal. Sports (MMSports)},
  pages 1--8, Lisbon, Port., Oct. 2022. ACM.

\bibitem{Vandeghen2022SemiSupervised}
Renaud Vandeghen, Anthony Cioppa, and Marc Van~Droogenbroeck.
\newblock Semi-supervised training to improve player and ball detection in
  soccer.
\newblock In {\em IEEE Int. Conf. Comput. Vis. Pattern Recognit. Work. (CVPRW),
  CVsports}, pages 3480--3489, New Orleans, LA, USA, Jun. 2022. Inst. Electr.
  Electron. Eng. (IEEE).

\bibitem{Vanderplaetse2020Improved}
Bastien Vanderplaetse and Stephane Dupont.
\newblock Improved soccer action spotting using both audio and video streams.
\newblock In {\em IEEE Int. Conf. Comput. Vis. Pattern Recognit. Work. (CVPRW),
  CVsports}, pages 3921--3931, Seattle, WA, USA, Jun. 2020. Inst. Electr.
  Electron. Eng. (IEEE).

\bibitem{Vats2020Event-arxiv}
Kanav Vats, Mehrnaz Fani, Pascale Walters, David~A. Clausi, and John Zelek.
\newblock Event detection in coarsely annotated sports videos via parallel
  multi receptive field {1D} convolutions.
\newblock {\em CoRR}, abs/2004.06172, 2020.

\bibitem{Vats2022IceHockey}
Kanav Vats, William McNally, Pascale Walters, David~A. Clausi, and John~S.
  Zelek.
\newblock Ice hockey player identification via transformers and weakly
  supervised learning.
\newblock In {\em IEEE/CVF Conf. Comput. Vis. Pattern Recognit. Work. (CVPRW)},
  pages 3450--3459, New Orleans, LA, USA, Jun. 2022. Inst. Electr. Electron.
  Eng. (IEEE).

\bibitem{yang2015multi}
Yi Yang, Zhigang Ma, Feiping Nie, Xiaojun Chang, and Alexander~G. Hauptmann.
\newblock Multi-class active learning by uncertainty sampling with diversity
  maximization.
\newblock {\em IJCV}, 113(2):113--127, Nov. 2014.

\bibitem{yoo2019learning}
Donggeun Yoo and In~So Kweon.
\newblock Learning loss for active learning.
\newblock In {\em IEEE/CVF Conf. Comput. Vis. Pattern Recognit. (CVPR)}, pages
  93--102, Long Beach, CA, USA, Jun. 2019. Inst. Electr. Electron. Eng. (IEEE).

\bibitem{Yu2018Comprehensive}
Junqing Yu, Aiping Lei, Zikai Song, Tingting Wang, Hengyou Cai, and Na Feng.
\newblock Comprehensive dataset of broadcast soccer videos.
\newblock In {\em IEEE Conf. Multimedia Inf. Process. Retr. (MIPR)}, pages
  418--423, Miami, FL, USA, Apr. 2018. Inst. Electr. Electron. Eng. (IEEE).

\bibitem{Zhu2022ATransformerbased}
He Zhu, Junwei Liang, Chengzhi Lin, Jun Zhang, and Jianming Hu.
\newblock A transformer-based system for action spotting in soccer videos.
\newblock In {\em Int. ACM Work. Multimedia Content Anal. Sports (MMSports)},
  pages 103--109, Lisbon, Port., Oct. 2022. ACM.

\end{thebibliography}
}

\end{document}